\documentclass{article}

\usepackage{arxiv}

\usepackage[utf8]{inputenc} 
\usepackage[T1]{fontenc}    
\usepackage{hyperref}       
\usepackage{url}            
\usepackage{booktabs}       
\usepackage{amsfonts}       
\usepackage{amsmath}        
\DeclareMathOperator*{\argmax}{arg\,max}
\usepackage{nicefrac}       
\usepackage{microtype}      
\usepackage{graphicx}
\usepackage[numbers]{natbib}
\usepackage{doi}
\usepackage{algorithm}
\usepackage{algorithmic}

\title{The Cognitive Companion: A Lightweight Parallel Monitoring Architecture for Detecting and Recovering from Reasoning Degradation in LLM Agents}

\author{
    Rafflesia Khan\\
    Software Developer, IBM\\
    Dublin, Ireland\\
    \texttt{rafflesiakhan.nw@gmail.com}
    \And
    Nafiul Islam Khan\\
    Diploma of Computer Science and Engineering\\
    Citi Polytechnic Institute\\
    Khulna, Bangladesh\\
    \texttt{earthkhan01@gmail.com}
}

\renewcommand{\shorttitle}{The Cognitive Companion}

\hypersetup{
hypertexnames=false,
pdftitle={The Cognitive Companion: A Lightweight Parallel Monitoring Architecture for Detecting and Recovering from Reasoning Degradation in LLM Agents},
pdfsubject={cs.AI, cs.CL},
pdfauthor={Rafflesia Khan and Nafiul Islam Khan},
pdfkeywords={Large Language Models, Agent Monitoring, Reasoning Degradation, Hidden State Probing},
}

\begin{document}
\maketitle

\begin{abstract}
Large language model (LLM) agents on multi-step tasks suffer reasoning degradation — looping, drift, stuck states — at rates up to 30\% on hard tasks. Current solutions include hard step limits (abrupt) or LLM-as-judge monitoring (10–15\% overhead per step). This paper introduces the \textbf{Cognitive Companion}, a parallel monitoring architecture with two implementations: an LLM-based Companion and a novel zero-overhead Probe-based Companion.

We report a three-batch feasibility study centered on Gemma 4 E4B, with an additional exploratory small-model analysis on Qwen 2.5 1.5B and Llama 3.2 1B. In our experiments, the \textbf{LLM-based Companion} reduced repetition on loop-prone tasks by 52–62\% with approximately 11\% overhead. The \textbf{Probe-based Companion}, trained on hidden states from layer 28, showed a mean effect size of +0.471 at \textbf{zero measured inference overhead}; its strongest probe result achieved cross-validated AUROC 0.840 on a small proxy-labeled dataset.

A key empirical finding is that \textbf{companion benefit appears task-type dependent}: companions are most helpful on loop-prone and open-ended tasks, while effects are neutral or negative on more structured tasks. Our small-model experiments also suggest a possible \textbf{scale boundary}: companions did not improve the measured quality proxy on 1B--1.5B models, even when interventions fired.

Overall, the paper should be read as a feasibility study rather than a definitive validation. The results provide encouraging evidence that sub-token monitoring may be useful, identify task-type sensitivity as a practical design constraint, and motivate selective companion activation as a promising direction for future work.
\end{abstract}


\section{Introduction}

Large language model (LLM) agents deployed on multi-step reasoning tasks exhibit systematic failure modes — repetitive loops, semantic drift, and stuck states — that closely mirror human cognitive degradation under load. Recent studies show that models with 1.5 billion parameters enter repetitive states on approximately 30\% of challenging reasoning tasks~\cite{pipis2025looping}, with this tendency paradoxically amplified in smaller distilled models.

Current mitigation strategies fall into three categories with significant limitations: (1) \textbf{Hard termination mechanisms} that abruptly halt execution, sacrificing productive reasoning; (2) \textbf{LLM-as-judge monitoring} requiring secondary inference passes, adding 10-15\% overhead~\cite{zheng2023judging}; and (3) \textbf{Token-level repetition penalties} that operate lexically without addressing semantic degradation.

This paper introduces the \textbf{Cognitive Companion}, a parallel monitoring architecture inspired by human cognitive support. Like a thinking partner who remains silent during productive work but intervenes constructively when obstacles arise, the Companion preserves agent autonomy while providing targeted assistance when degradation is detected.

We conducted three experimental batches: (1) \textbf{E1}: Initial validation with Gemma 4 E4B using LLM-based monitoring; (2) \textbf{E2}: Comprehensive six-task evaluation comparing baseline, LLM Companion, and Probe Companion approaches; (3) \textbf{E3}: Cross-model validation on small models (Qwen 2.5 1.5B, Llama 3.2 1B) to identify scale boundaries.

Our investigation addresses: (1) Can reasoning degradation be detected through zero-overhead mechanisms using internal model representations? (2) When do companions help versus harm performance? (3) What is the minimum viable model scale for companion effectiveness?

\subsection{Contributions}

This feasibility study makes five key contributions:

\begin{enumerate}
\item \textbf{Architectural Framework}: We formalize the Cognitive Companion architecture with three intervention modes (whisper, surface, autonomous) and provide a complete implementation framework.

\item \textbf{Probe-based Monitoring Prototype}: We develop a Probe-based Companion using hidden states from layer 28 and show preliminary evidence that hidden-state monitoring can improve in-domain outcomes with zero measured computational overhead. In our current experiments, the Probe-based Companion achieved mean effect size +0.471, while the strongest probe training run reached AUROC 0.840 on a small proxy-labeled dataset.

\item \textbf{Task-Type Sensitivity Discovery}: We show that companion benefit is task-dependent in our current setup: improvements are concentrated on loop-prone and open-ended tasks, while effects are neutral-to-harmful on structured tasks (database decisions, algorithms). This finding suggests selective rather than universal deployment.

\item \textbf{Exploratory Scale Boundary Evidence}: We provide negative exploratory evidence that models below ~3B parameters may struggle to act on companion guidance, observing no improvement in the measured quality proxy across Qwen 2.5 1.5B and Llama 3.2 1B despite successful intervention firing.

\item \textbf{Transparent Evaluation Framework}: We provide a multi-condition evaluation framework combining Jaccard repetition, proxy quality scoring, effect size reporting, and explicit disclosure of methodological limitations including proxy labels, self-referential judging, and small-sample uncertainty.
\end{enumerate}

\subsection{Scope and Limitations}

This work represents a preliminary feasibility study conducted primarily with Gemma 4 E4B on a small set of reasoning tasks. We acknowledge several methodological constraints that limit the generalizability of the current results: the probe training dataset contains only 35 proxy-labeled examples, evaluation relies partially on self-referential judging, the small-model study uses a simpler quality proxy than the main Gemma evaluation, and experiments are conducted within a narrow model and task range. These limitations are addressed through transparent reporting and a future work roadmap rather than being treated as resolved.

Despite these constraints, the results provide encouraging evidence for lightweight cognitive monitoring and a concrete methodological foundation for more rigorous follow-up evaluation. The zero-overhead Probe-based approach is best interpreted here as a promising proof of concept rather than a production-ready monitoring solution.

\section{Background and Related Work}

\subsection{Reasoning Degradation in LLM Agents}

The phenomenon of LLM agents entering degenerative states during extended reasoning has been characterized from multiple perspectives. At the token level, repetitive generation patterns are well understood: models assign increasing probability to recently generated tokens, creating self-reinforcing cycles that can lead to deterministic loops~\cite{holtzman2020curious}. However, this work focuses on \textbf{semantic-level degradation}, where agents may produce lexically varied outputs that nonetheless revisit identical conceptual territory without advancing toward task objectives.

Pipis et al.~\cite{pipis2025looping} provide the most systematic characterization of semantic looping in reasoning models, identifying two primary mechanisms: (1) \textit{risk aversion during training}, where models more readily acquire cyclic action patterns than complex progress-making behaviors, and (2) an \textit{inductive bias in transformer architectures} toward temporally correlated error patterns. Their critical finding that smaller distilled models exhibit significantly higher looping rates than their teacher counterparts directly motivates the need for external monitoring mechanisms during deployment.

The practical impact of reasoning degradation extends beyond individual task failures. Looping agents consume computational resources inefficiently, degrade the signal-to-noise ratio in their context windows, and may mislead human operators who assume continued generation indicates productive work. These factors combine to create substantial operational costs in production environments where agent reliability is paramount.

\subsection{Hidden State Probing and the Semantic Capacity Asymmetry Hypothesis}

The application of linear classifiers trained on transformer hidden states to detect latent properties has established a rich methodological foundation. Alain and Bengio~\cite{alain2017understanding} pioneered the systematic study of intermediate layer representations, demonstrating that simple logistic regression classifiers can accurately identify syntactic and semantic properties from frozen activations. This framework has since been extended to critical applications including hallucination detection~\cite{burns2023discovering}, truthfulness assessment~\cite{li2023inference}, and confidence calibration~\cite{kuhn2023semantic}.

Central to our Probe-based Companion design is the \textbf{Semantic Capacity Asymmetry Hypothesis}, formalized in the INSPECTOR framework~\cite{inspector2026}. This hypothesis posits that evaluating a reasoning step requires substantially less semantic capacity than generating that step, implying that model hidden states encode rich evaluative signals even when generated outputs fail to articulate them explicitly. This asymmetry provides the theoretical foundation for detecting reasoning quality from internal representations without requiring full model inference.

Empirical support for this hypothesis comes from multiple sources. The ICR Probe work~\cite{burns2023discovering} demonstrates that hallucination signals are strongest in deeper transformer layers (layers 29-41 in their study), while confidence calibration research consistently finds that middle-to-late layers encode the most reliable quality indicators~\cite{kuhn2023semantic}. Our results extend this pattern to reasoning degradation, with layer 28 of Gemma 4 E4B yielding optimal cross-validated performance among tested layers.

\subsection{Production Agent Monitoring}

Current production agent frameworks implement loop detection through predominantly heuristic approaches. LangGraph employs simple iteration counting with a configurable \texttt{recursion\_limit} parameter (default 25 steps)~\cite{langgraph2024}. AutoGen provides composable termination conditions based on message count, token budget, and basic pattern matching~\cite{wu2023autogen}. OpenHands represents the current state-of-the-art with a \texttt{StuckDetector} module that identifies five distinct stagnation patterns through entirely rule-based heuristics~\cite{opendevin2024}.

Notably, none of these widely-adopted frameworks incorporate learned representations or hidden state analysis for semantic-level degradation detection. This represents a significant gap given that rule-based approaches cannot capture the subtle semantic patterns that characterize reasoning loops in natural language generation.

Recent research has explored more sophisticated monitoring approaches. SpecRA~\cite{specra2025} applies randomized Fast Fourier Transform analysis to token sequences for detecting periodic patterns at O(n log n) computational cost. ERGO~\cite{ergo2025} monitors Shannon entropy over next-token probability distributions as a real-time degradation signal. While promising, these approaches operate primarily at the token or probability level rather than the semantic level targeted by our work.

The Cognitive Companion architecture complements these existing approaches by incorporating semantic-level analysis through hidden state monitoring while maintaining the intervention and recovery capabilities necessary for practical deployment.

\subsection{Self-Correction Limitations in Small Language Models}

A fundamental constraint influencing Cognitive Companion design is the established difficulty of self-correction in smaller LLMs. Huang et al.~\cite{huang2024large} demonstrate that models below approximately 13 billion parameters struggle to reliably self-correct reasoning errors through prompting alone. While recent work such as STaSC~\cite{stasc2025} has shown that self-correction capabilities can be trained into smaller models through iterative refinement on filtered examples, this approach requires model fine-tuning and substantial training data.

The Cognitive Companion sidesteps this limitation by providing \textbf{external correction guidance} rather than relying on intrinsic model self-awareness. This design choice is particularly important for resource-constrained deployments where smaller, more efficient models are preferred but lack the scale necessary for reliable self-evaluation. By offering targeted reorientation from an external monitoring system, the Companion enables effective error recovery without requiring modification to the primary agent's parameters or training.

\section{Problem Formulation}

\subsection{Agent Execution Model}

We model the primary agent's operation as a sequential reasoning process over a discrete time horizon. Let $T$ represent the initial task description provided to the agent. The agent's cumulative step history after $t$ reasoning steps is denoted as:

$$H_t = \{(p_1, r_1), (p_2, r_2), \ldots, (p_t, r_t)\}$$

where $p_i$ represents the prompt constructed for step $i$ and $r_i$ denotes the corresponding response generated by the primary language model $M$. At each subsequent step, the agent constructs its prompt from the task context and recent history, generating the next response according to:

$$r_{t+1} = M(T, H_t, G_t)$$

where $G_t$ represents the set of guidance injections provided by the Cognitive Companion up to step $t$. In baseline experimental conditions, $G_t$ is consistently empty, representing autonomous agent operation without external intervention.

\subsection{Cognitive State Classification}

To enable systematic monitoring and intervention, we define four distinct cognitive states that characterize agent behavior at any given reasoning step:

\begin{itemize}
\item \textbf{ON\_TRACK}: The agent demonstrates genuine progress toward task completion, with each step introducing novel reasoning, evidence, or conclusions not present in prior steps.

\item \textbf{LOOPING}: The agent exhibits repetitive reasoning patterns, revisiting prior arguments or conceptual territory without substantive advancement toward the task objective.

\item \textbf{DRIFTING}: The agent's reasoning trajectory diverges from the original task description, with declining topical relevance across successive steps.

\item \textbf{STUCK}: The agent fails to make meaningful progress and exhibits indicators of declining reasoning quality, such as shortened responses or increased expressions of uncertainty.
\end{itemize}

For the Probe-based Companion implementation, we collapse this taxonomy into a binary classification problem to simplify learning and improve reliability given limited training data:

$$y_t \in \{0, 1\} \text{ where } 0 = \text{ON\_TRACK}, \; 1 = \text{DEGRADED}$$

where DEGRADED encompasses the union of LOOPING, DRIFTING, and STUCK states.

\subsection{Companion Objective}

The Cognitive Companion's overarching objective is to implement a function $C$ that, given recent agent history and optionally the primary model's internal states, determines the necessity and nature of intervention:

$$(intervene_t, guidance_t) = C(H_{t-k:t}, \Phi_t, \theta)$$

where $H_{t-k:t}$ denotes the $k$ most recent steps in the agent's reasoning history, $\Phi_t$ represents hidden state representations extracted from the primary model at step $t$, and $\theta$ encompasses the Companion's learned parameters.

The Companion's performance is optimized by minimizing a weighted loss function that balances detection accuracy with computational efficiency:

$$\mathcal{L} = \alpha \cdot \mathbf{1}[\text{degraded}_t=1, \text{intervene}_t=0] + \beta \cdot \mathbf{1}[\text{degraded}_t=0, \text{intervene}_t=1] + \gamma \cdot \text{overhead}(C)$$

The three terms penalize: (1) \textbf{missed interventions} (false negatives) where degradation occurs but the Companion fails to act; (2) \textbf{unnecessary interventions} (false positives) where the Companion interrupts productive reasoning; and (3) \textbf{computational overhead} imposed by the monitoring system. The weighting coefficients $\alpha > \beta > \gamma$ reflect our priority ordering: missing genuine degradation is the most harmful outcome, while computational overhead is the least costly concern provided the core monitoring functionality is preserved.

\section{Architecture}

\subsection{System Overview}

The Cognitive Companion architecture consists of three interconnected components designed for semi-parallel operation with minimal disruption to primary agent reasoning: the \textbf{Primary Agent}, the \textbf{Companion Observer}, and the \textbf{Intervention Handler}. This modular design enables flexible deployment across different agent frameworks while maintaining clear separation of concerns.

The Primary Agent executes reasoning steps autonomously, generating responses according to its standard inference process. Concurrently, the Companion Observer periodically samples both the agent's textual output and, crucially, its internal hidden state representations. When reasoning degradation is detected, the Intervention Handler injects targeted guidance into the agent's context, facilitating recovery while preserving the agent's sense of autonomous operation.

\begin{figure}[t]
\centering
\includegraphics[width=0.75\linewidth]{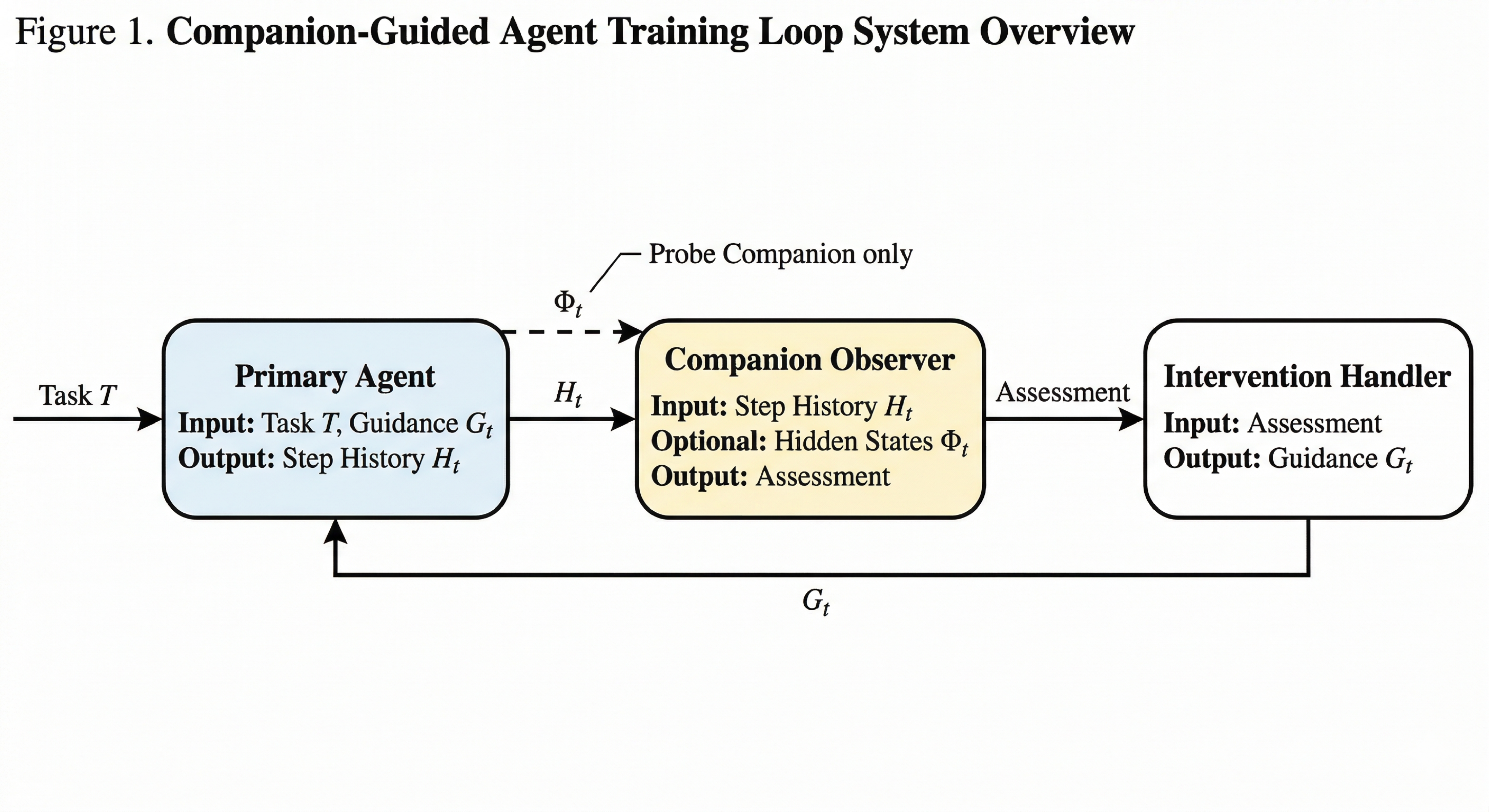}
\caption{System overview showing the three-component Cognitive Companion architecture. The Primary Agent executes reasoning steps while the Companion Observer monitors cognitive state and the Intervention Handler injects guidance when degradation is detected.}
\label{fig:system_overview}
\end{figure}

Figure~\ref{fig:system_overview} summarizes this three-component architecture and highlights the information flow from task input through observation to intervention.

The \textbf{Primary Agent} executes the assigned task through iterative reasoning steps, maintaining a step history $H_t = \{r_1, r_2, \ldots, r_t\}$ where each $r_i$ represents a reasoning response. The \textbf{Companion Observer} monitors the agent's cognitive state through either periodic LLM-based assessment or continuous hidden state analysis. The \textbf{Intervention Handler} translates detected cognitive issues into constructive guidance $G_t$ that is injected into the agent's context when necessary.

A critical design principle is that the Primary Agent remains architecturally unchanged and unaware of the Companion's explicit presence. This preserves the agent's native reasoning patterns while ensuring that observed improvements are directly attributable to Companion interventions.

\subsection{LLM-based Companion (LC)}

\begin{figure}[htbp]
    \centering
    \includegraphics[width=0.75\linewidth,height=0.58\textheight,keepaspectratio]{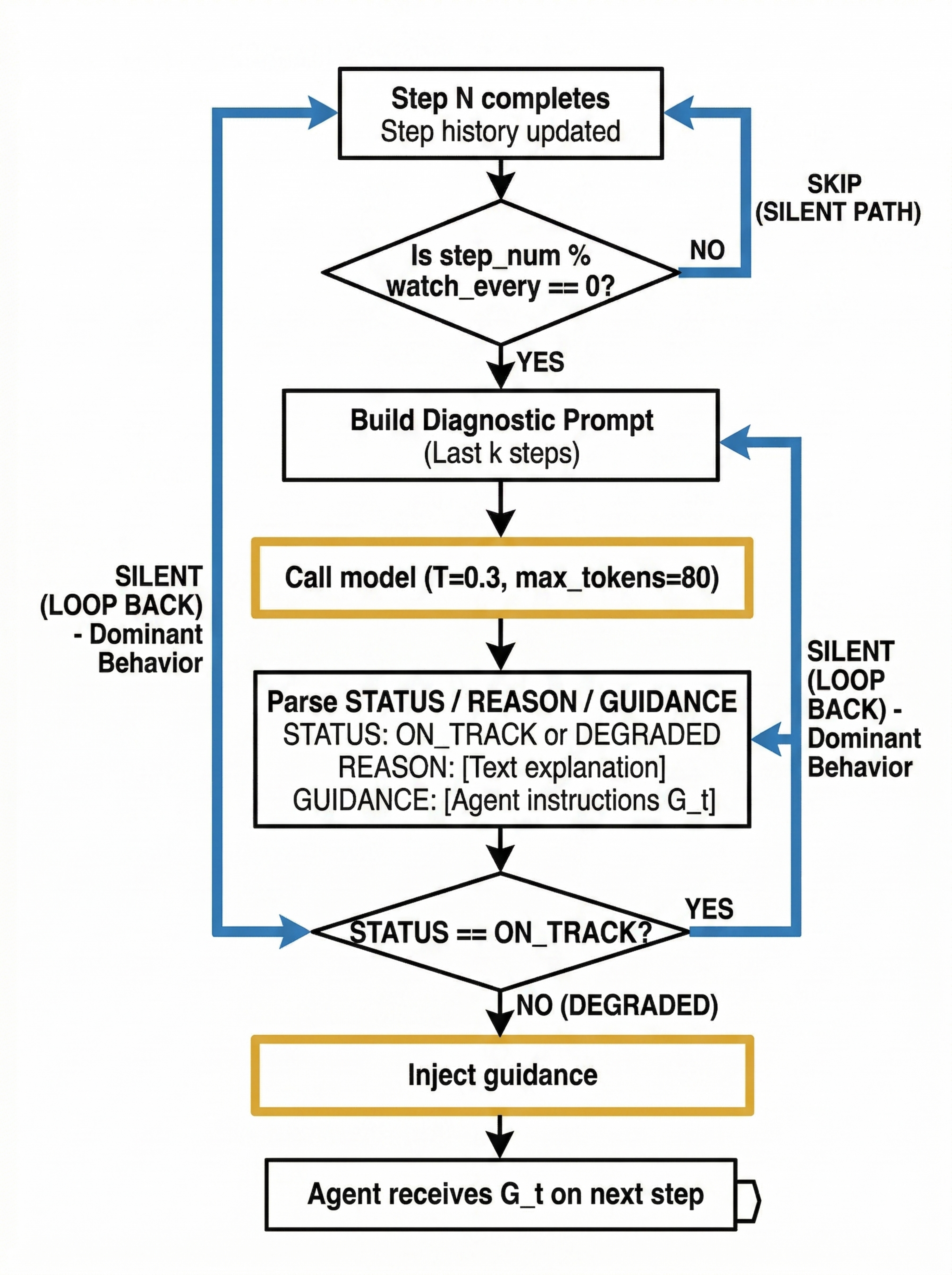}
    \caption{LLM Companion decision cycle showing periodic assessment with structured prompts and selective intervention.}
    \label{fig:llm_companion_flow}
\end{figure}

The LLM-based Companion leverages the primary agent model itself for periodic cognitive state assessment. Figure~\ref{fig:llm_companion_flow} details the complete decision cycle.

At predefined intervals (controlled by \texttt{watch\_every}), the Companion constructs a diagnostic prompt requesting structured assessment:

\begin{verbatim}
Recent agent steps: [step history]

Assess the agent's cognitive state:
STATUS: [LOOPING/DRIFTING/STUCK/ON_TRACK]  
REASON: [one sentence explanation]
GUIDANCE: [intervention text or NONE]
\end{verbatim}

The Companion operates with temperature=0.3 and 80-token maximum to ensure parsing reliability. Periodic assessment using structured prompts achieves compute overhead of ~10–12\% with \texttt{watch\_every=2}.

\subsection{Probe-based Companion (PC)}

\begin{figure}[htbp]
    \centering
    \includegraphics[width=0.95\linewidth]{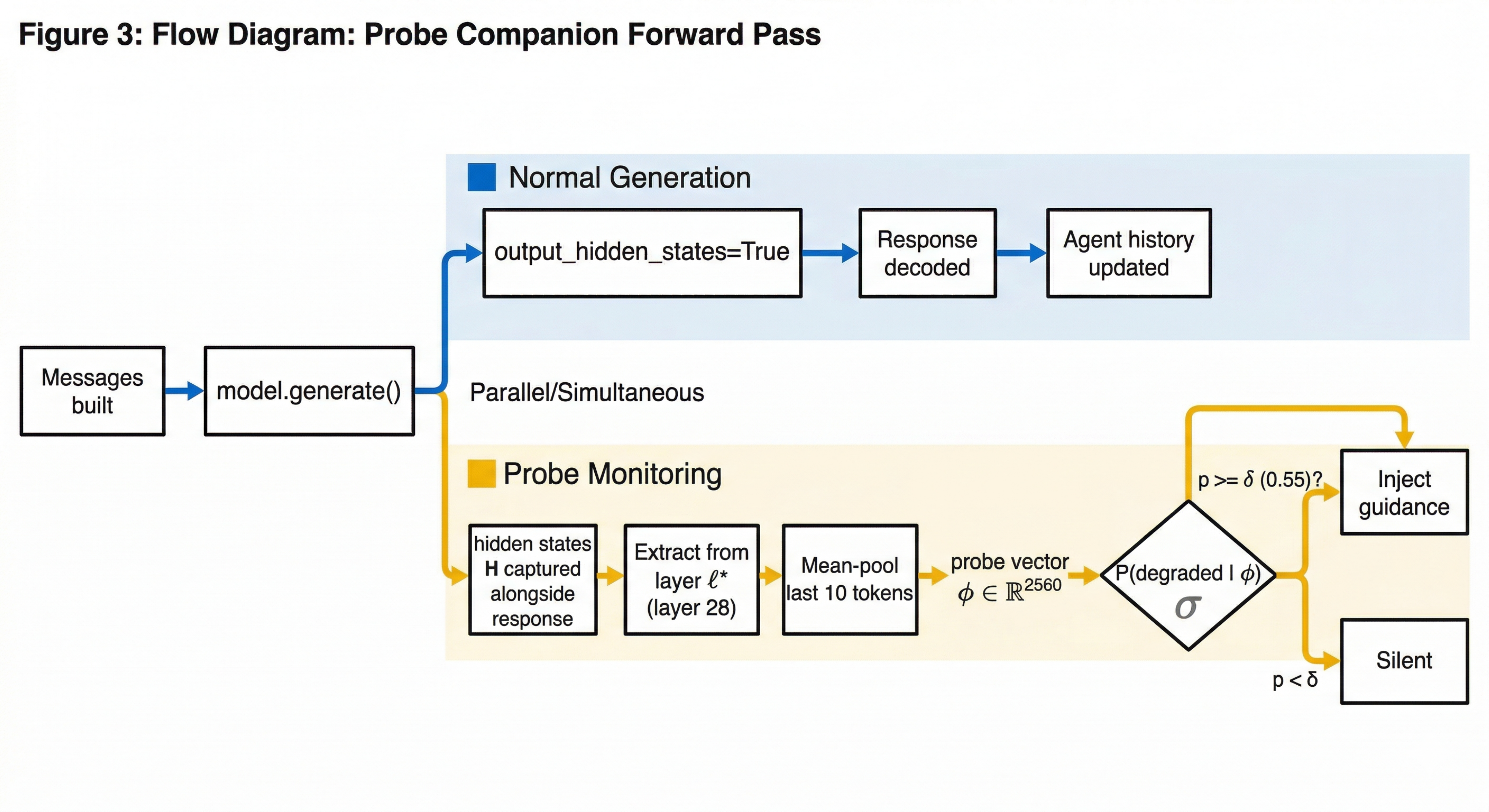}
    \caption{Probe Companion forward pass showing hidden state extraction during existing generation with zero additional model calls.}
    \label{fig:probe_companion_flow}
\end{figure}

The Probe-based Companion represents our core architectural innovation, achieving degradation detection through hidden state analysis during the primary model's existing forward pass. Figure~\ref{fig:probe_companion_flow} shows the zero-overhead extraction process.

\subsubsection{Hidden State Extraction}

Gemma 4 E4B employs a standard decoder-only transformer architecture with $L=42$ layers, each producing hidden state tensors of shape $[\text{batch}, \text{seq\_len}, d_{\text{model}}]$ where $d_{\text{model}}=2560$. Hidden states are extracted by enabling \texttt{output\_hidden\_states=True} during the primary model's generation process, incurring minimal memory overhead.

For each reasoning step, we extract probe features by mean-pooling the final-position hidden states across the last $w$ generated tokens:

$$\phi_{t,\ell} = \frac{1}{w} \sum_{i=n-w+1}^{n} h_{i,\ell}[-1, :]$$

where $h_{i,\ell}[-1, :]$ denotes the final-position hidden state at layer $\ell$ for the $i$-th generated token, $n$ is the total sequence length, and $w=10$ represents a configurable window parameter. This mean-pooling strategy captures semantic information from the model's most recent reasoning while providing robust features for classification.

\begin{figure}[htbp]
    \centering
    \includegraphics[width=0.90\linewidth,height=0.58\textheight,keepaspectratio]{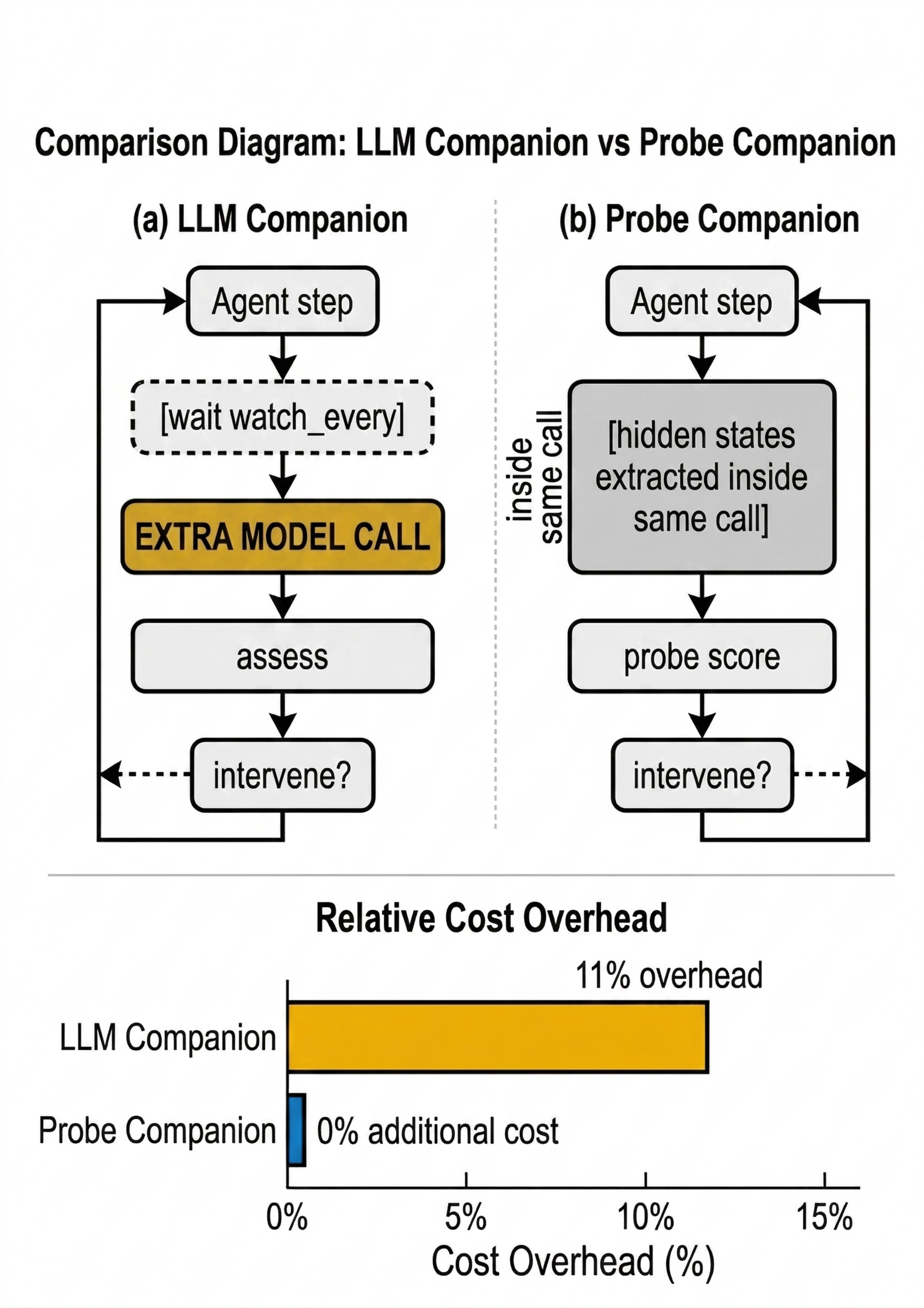}
    \caption{Architectural comparison showing the fundamental overhead difference between LLM-based and Probe-based monitoring approaches.}
    \label{fig:companion_comparison}
\end{figure}

Figure~\ref{fig:companion_comparison} contrasts the LLM-based and Probe-based implementations at the architectural level, emphasizing that the probe pathway reuses the primary forward pass rather than requiring a second model invocation.

\subsubsection{Probe Training Methodology}

We train binary logistic regression classifiers $f_\ell: \mathbb{R}^{2560} \to [0,1]$ independently at candidate layers $\Lambda = \{20, 28, 34, 40\}$:

\begin{align}
P(y_t = 1 | \phi_{t,\ell}) &= \sigma(w_\ell^T \cdot \phi_{t,\ell} + b_\ell) \\
\ell^* &= \argmax_{\ell \in \Lambda} \text{AUROC}_{\text{CV}}(f_\ell)
\end{align}

Hidden state extraction from existing forward pass achieves zero measured inference overhead. Linear classifier trained on mean-pooled final tokens with L2 regularization and balanced class weighting.

\subsection{Intervention Modes}

\begin{figure}[htbp]
    \centering
    \includegraphics[width=0.85\linewidth]{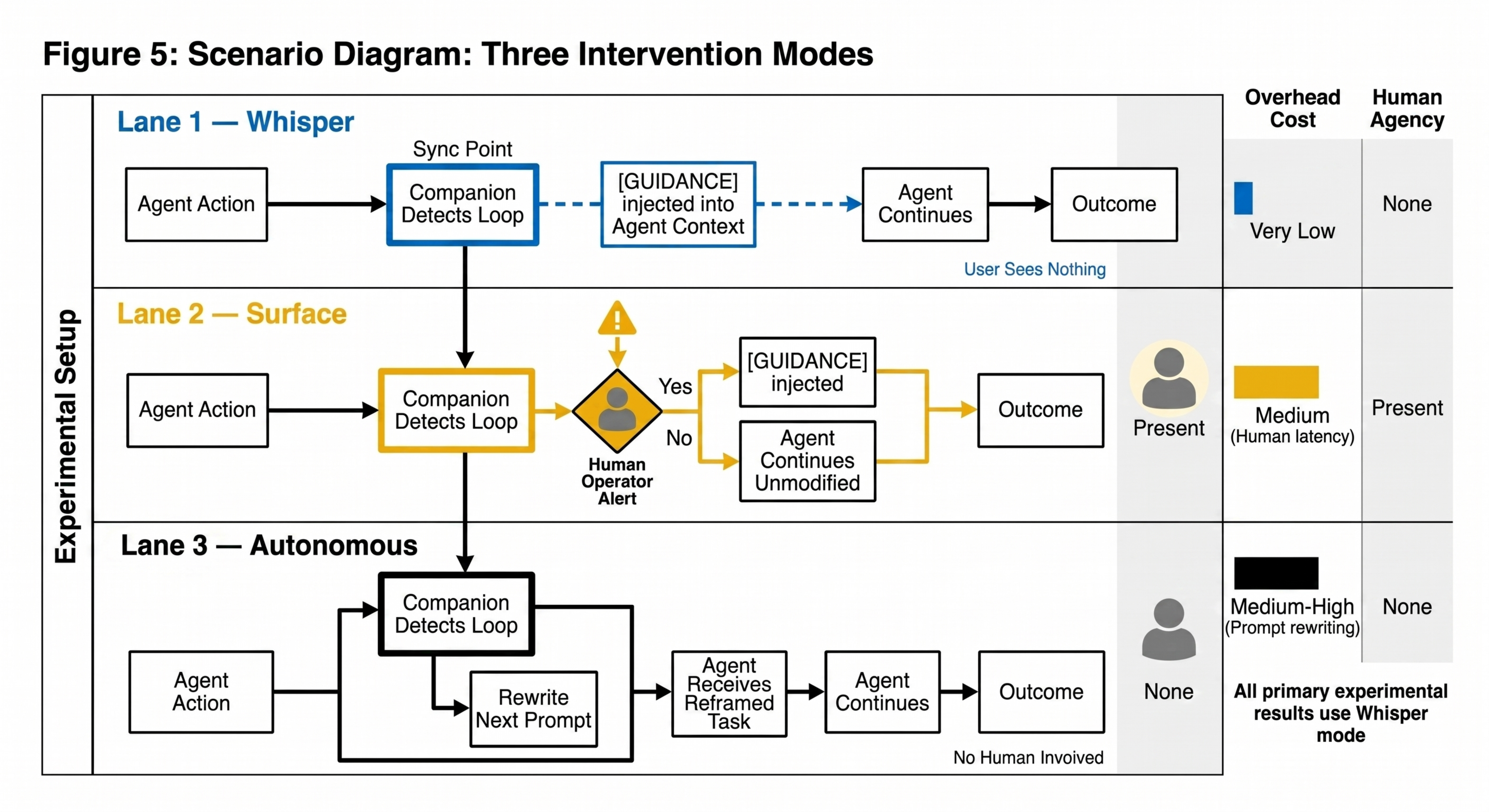}
    \caption{Three intervention modes supporting different deployment scenarios from fully automated to human-in-the-loop monitoring.}
    \label{fig:intervention_modes}
\end{figure}

The architecture supports three distinct intervention modes for different deployment scenarios. Figure~\ref{fig:intervention_modes} illustrates each approach.

\textbf{Whisper Mode}: Silent guidance injection into agent context (used in all experiments). \textbf{Surface Mode}: Human operator receives alerts and decides on intervention. \textbf{Autonomous Mode}: Direct task reframing without human involvement.

\subsubsection{Deployment and Overhead Analysis}

The Probe-based Companion achieves \textbf{zero measured inference overhead}: (1) Hidden state extraction occurs during existing forward pass; (2) Probe classification involves single matrix multiplication; (3) No additional LLM calls required.

\section{Experimental Design}

\subsection{Experimental Setup}

We conducted three experimental batches addressing different research questions. Table~\ref{tab:experiment_overview} summarizes the complete experimental design.

\begin{table}[htbp]
\centering
\small
\begin{tabular}{|l|l|l|l|c|p{3.7cm}|}
\hline
\textbf{Batch} & \textbf{Notebook} & \textbf{Model} & \textbf{Conditions} & \textbf{Tasks} & \textbf{Research Question} \\
\hline
E1 & v4 & Gemma 4 E4B & Baseline, LLM & 1 (Liar Paradox) & Does companion work at all? \\
E2 & v5 & Gemma 4 E4B & Baseline, LLM, Probe & 6 (diverse types) & When does it work/not work? \\
E3 & small models & Qwen 1.5B, Llama 1B & Baseline, LLM & 3 tasks & Minimum viable scale? \\
\hline
\end{tabular}
\caption{Three experimental batches with progressive scope and complexity. E1 probes feasibility, E2 explores task-type dependencies, and E3 investigates possible scale boundaries.}
\label{tab:experiment_overview}
\end{table}

All primary experiments used Gemma 4 E4B (\texttt{google/gemma-4-E4B-it}) with standard sampling parameters (temperature=1.0, top\_p=0.95, top\_k=64). The main v5 evaluation used 6 reasoning steps per run, companion checks every 2 steps, and 3 runs per condition. This represents a \textbf{preliminary feasibility study} with acknowledged methodological limitations addressed in future work.

\subsection{Implementation Details and Reproducibility}

The primary agent and LLM-based Companion use the same base model. The agent generates with temperature 1.0, top\_p 0.95, and top\_k 64. The LLM-based Companion assesses the three most recent steps and emits structured fields \texttt{STATUS}, \texttt{REASON}, and \texttt{GUIDANCE}, which are parsed into intervention decisions.

The Probe-based Companion trains a separate logistic regression classifier at each candidate layer in $\Lambda = \{20, 28, 34, 40\}$ using balanced class weighting, \texttt{random\_state=42}, and \texttt{max\_iter=1000}. Probe intervention uses a fixed degradation threshold of 0.55 and checks every 2 agent steps. Probe training labels are obtained by running the LLM-based Companion at every step and collapsing \texttt{LOOPING}, \texttt{DRIFTING}, and \texttt{STUCK} into a binary \texttt{DEGRADED} label.

Two methodological details are important for interpreting the results. First, the current v5 probe experiment is \textit{in-domain}: the probe is trained on a subset of tasks drawn from the same task bank later used for evaluation, so the resulting probe gains should not be interpreted as strong out-of-domain generalization evidence. Second, the E3 small-model study uses a simpler quality proxy than the Gemma-based E1/E2 experiments, so those quality numbers support negative exploratory evidence rather than direct apples-to-apples comparison.

\subsection{Task Design}

We categorized six reasoning tasks by their expected failure modes. Table~\ref{tab:task_taxonomy} provides the complete taxonomy.

\begin{table}[htbp]
\centering
\begin{tabular}{|l|l|l|l|}
\hline
\textbf{Category} & \textbf{Tasks} & \textbf{Defining Property} & \textbf{Expected Failure Mode} \\
\hline
\textbf{Loop-prone} & Liar Paradox, Ship of Theseus & No unique resolution & Cycles between logical positions \\
\textbf{Drift-prone} & Consciousness, Startup design & Open-ended scope & Wanders into tangents \\
\textbf{Structured} & DB decision, Algorithm design & Clear deliverable & Has natural stopping criterion \\
\hline
\end{tabular}
\caption{Task taxonomy showing three categories with different expected companion effectiveness profiles.}
\label{tab:task_taxonomy}
\end{table}

Tasks were selected to elicit specific failure modes while representing complex reasoning problems that production agents encounter. The category-based design enables systematic analysis of task-type effects on companion performance.

\subsection{Evaluation Metrics and Effect Size Reporting}

We employ Cohen's d as the primary effect size metric with the convention: |d| < 0.2 = negligible, 0.2–0.5 = small, 0.5–0.8 = medium, >0.8 = large effects.

\subsubsection{Repetition Measurement}

Reasoning degradation, particularly looping behavior, was quantified using word-level Jaccard similarity between consecutive reasoning steps:

$$J(r_t, r_{t+1}) = \frac{|W(r_t) \cap W(r_{t+1})|}{|W(r_t) \cup W(r_{t+1})|}$$

where $W(r)$ denotes the set of unique tokens in response $r$. We computed both mean and maximum Jaccard similarity across all consecutive step pairs. Jaccard similarity was selected over semantic similarity measures for its model-independence and direct interpretability as lexical overlap, providing a conservative lower bound on true semantic repetition.

\subsubsection{Quality Assessment}

Reasoning quality in E1 and E2 was evaluated using an adapted MT-Bench-style framework where each step was scored on three dimensions: Relevance (alignment with task objective), Progress (advancement toward completion), and Coherence (logical consistency). Scores were assigned on a 0-10 scale and aggregated into composite quality measures.

\textbf{Important Limitation}: For E1 and E2, quality assessment was performed by the primary model itself, introducing potential circular bias. For E3, the small-model notebook uses a simpler heuristic quality proxy based primarily on response length and basic completeness heuristics. The quality metrics therefore serve as approximate within-experiment indicators, not a single uniform gold-standard measure across all batches.

\begin{table}[htbp]
\centering
\small
\begin{tabular}{|p{4cm}|p{3cm}|p{3cm}|}
\hline
\textbf{Limitation} & \textbf{Impact} & \textbf{Future Work?} \\
\hline
Single model (Gemma 4 E4B) & Limits generalizability & Yes — cross-model probe \\
Probe dataset: 3 DEGRADED in v5 (9.4\%) & CV AUC = NaN & Yes — target 200/class \\
Self-referential judging & Potential circular bias & Yes — external judge \\
No significance testing & Effect sizes are estimates & Yes — multi-run framework \\
Probe NaN AUC in v5 & Explicitly disclosed & Acknowledged \\
\hline
\end{tabular}
\caption{Known limitations with impact assessment and future work mitigation plan.}
\label{tab:limitations}
\end{table}

\subsubsection{Computational Overhead}

Overhead was measured as wall-clock execution time consumed by Companion operations relative to total execution time:

$$\text{overhead} = \frac{t_{\text{companion}}}{t_{\text{agent}} + t_{\text{companion}}} \times 100\%$$

\subsection{Probe Training Data Collection}

Training data for the Probe-based Companion was collected by executing the LLM-based Companion for 8 steps per task and treating its STATUS assessments as proxy labels. In the strongest reported run, this process yielded 35 labeled examples with the distribution: ON\_TRACK (25 examples, 71.4\%), DEGRADED (10 examples, 28.6\%).

\textbf{Critical Limitation}: This dataset size is insufficient for robust generalization. In addition, the current v5 setup trains on tasks that overlap with the broader evaluation bank, making the probe results best interpreted as in-domain proof-of-concept validation rather than held-out task generalization. Production deployment would require substantially larger datasets, independent labeling or external adjudication, and clean train/validation/test splits across task families and model architectures.

\subsection{Threats to Validity}

Several threats to validity materially shape the interpretation of this study. \textbf{Proxy-label risk}: Probe supervision is derived from LLM Companion assessments rather than independent human or external-model annotations. \textbf{Self-judging risk}: E1 and E2 quality scores are produced by the same family of model used for generation, creating possible circularity and observer bias. \textbf{Small-sample risk}: The probe dataset is small, class-imbalanced, and unstable across collections, as evidenced by the v5 run with only 3 DEGRADED examples and undefined cross-validation AUROC. \textbf{In-domain evaluation risk}: The v5 probe experiment uses training tasks drawn from the same task bank as evaluation, limiting claims about generalization. \textbf{Metric inconsistency across batches}: E3 uses a simpler quality proxy than E1/E2, so cross-batch comparisons should be read qualitatively. These threats do not negate the feasibility results, but they substantially narrow the strength of the claims that can be made.

\section{Results}

\subsection{Experiment 1: Initial Validation (E1)}
\textit{From v4 notebook, Liar Paradox, LLM Companion, two sessions}

Experiment 1 provided initial validation of the LLM-based Companion approach using the Liar Paradox task across two independent sessions. Table~\ref{tab:e1_results} summarizes the quantitative improvements.

\begin{figure}[htbp]
    \centering
    \includegraphics[width=0.65\linewidth]{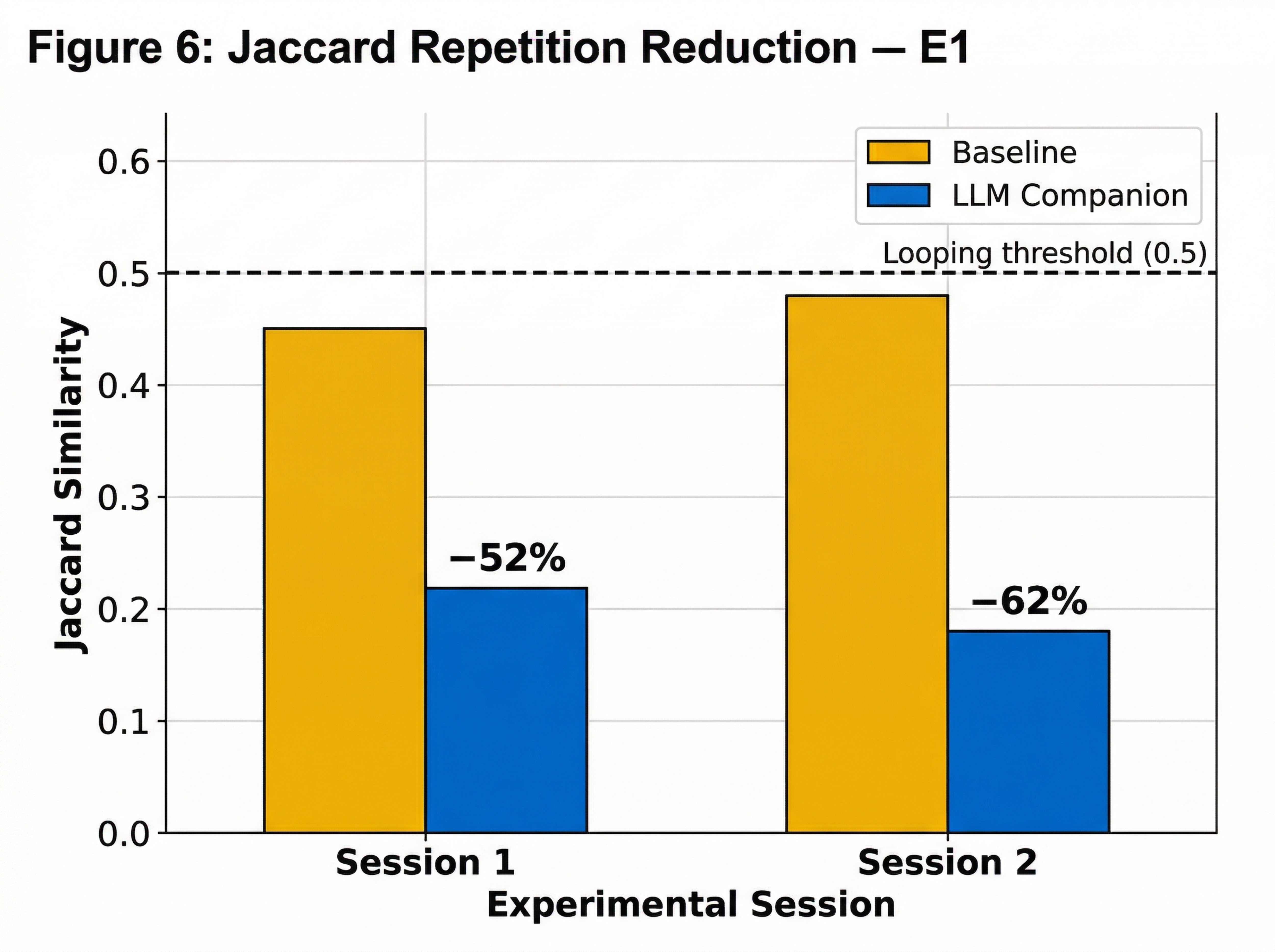}
\caption{Jaccard repetition reduction across two independent sessions in the initial feasibility study.}
    \label{fig:e1_repetition}
\end{figure}

\begin{figure}[htbp]
    \centering
    \includegraphics[width=0.85\linewidth]{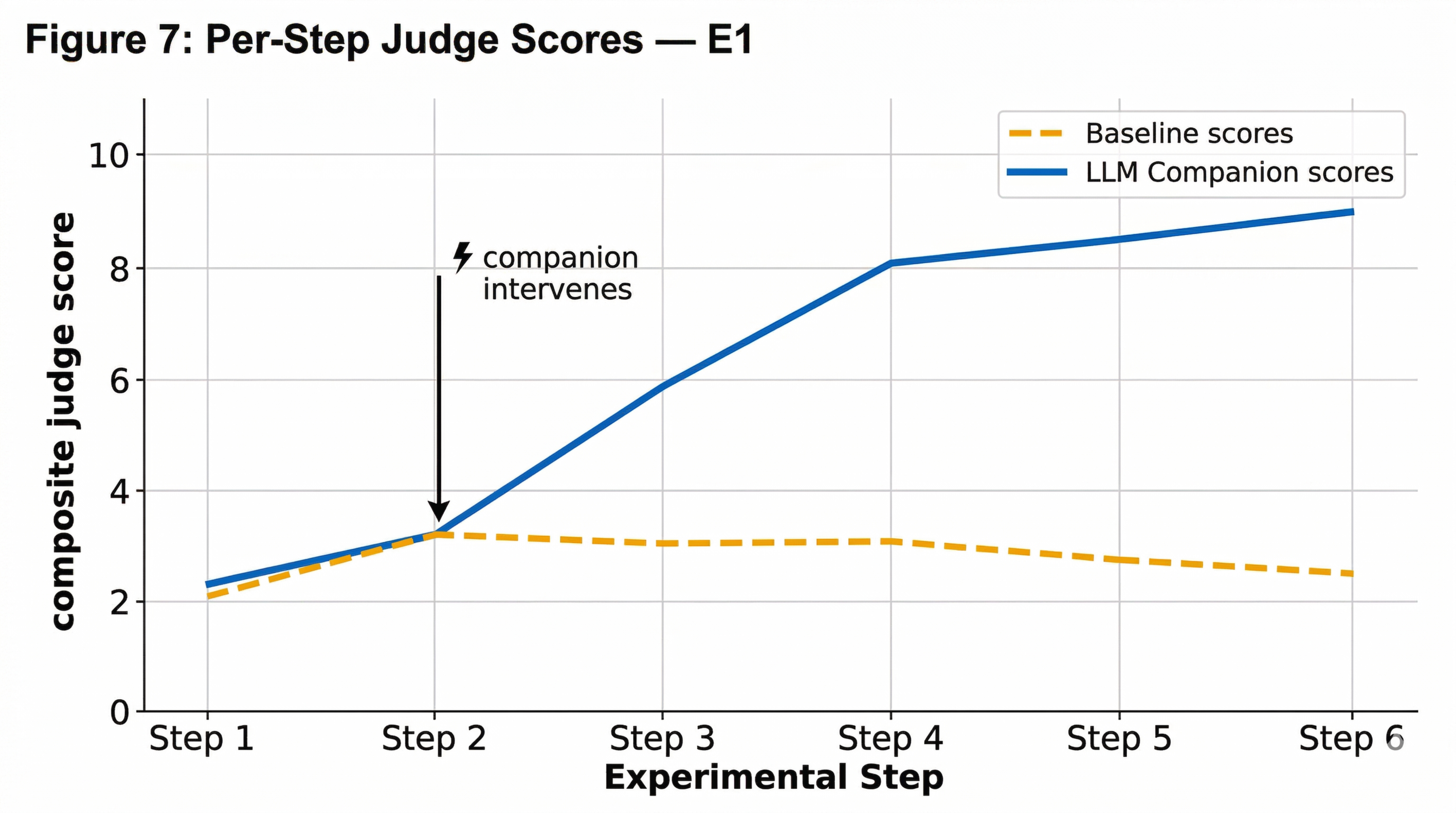}
\caption{Per-step proxy quality progression in an initial session, with improvement following companion intervention at step 2.}
    \label{fig:e1_progression}
\end{figure}

Figure~\ref{fig:e1_repetition} visualizes the reduction in step-to-step repetition across the two initial sessions, while Figure~\ref{fig:e1_progression} shows the corresponding per-step proxy quality trajectory around the intervention point.

\begin{table}[htbp]
\centering
\begin{tabular}{|l|c|c|c|}
\hline
\textbf{Metric} & \textbf{Baseline} & \textbf{+ LLM Companion} & \textbf{$\Delta$} \\
\hline
Mean judge score (/10) & 8.8 & 9.3 & \textbf{+0.5} \\
Final step score (/10) & 8.3 & 9.0 & +0.7 \\
Repetition mean (Jaccard) & 0.367 & 0.176 & \textbf{-52\%} \\
Repetition max & 0.480 & 0.329 & -32\% \\
Length trend & -16.09 & -1.97 & stabilised \\
Task completed & YES & YES & — \\
Overhead & 0\% & 10.2\% & — \\
Interventions fired & 0 & 1 (Step 2) & — \\
\hline
\end{tabular}
\caption{E1: LLM Companion vs Baseline on the Liar Paradox task, showing improved repetition reduction and proxy quality in two initial sessions.}
\label{tab:e1_results}
\end{table}

\subsection{Experiment 2: Multi-Task Multi-Condition Evaluation (E2)}
\textit{From v5 notebook, 6 tasks × 3 conditions × 3 runs}

Experiment 2 provided a broader in-domain evaluation across six tasks with three conditions (baseline, LLM Companion, Probe Companion). Tables~\ref{tab:e2_per_task} and~\ref{tab:e2_aggregate} present the complete results. Because the probe was trained on tasks drawn from the same task bank, these results should be read as preliminary evidence of usefulness within the current experimental distribution rather than as a strong claim of task-general probe transfer.

\begin{figure}[htbp]
    \centering
    \includegraphics[width=0.85\linewidth]{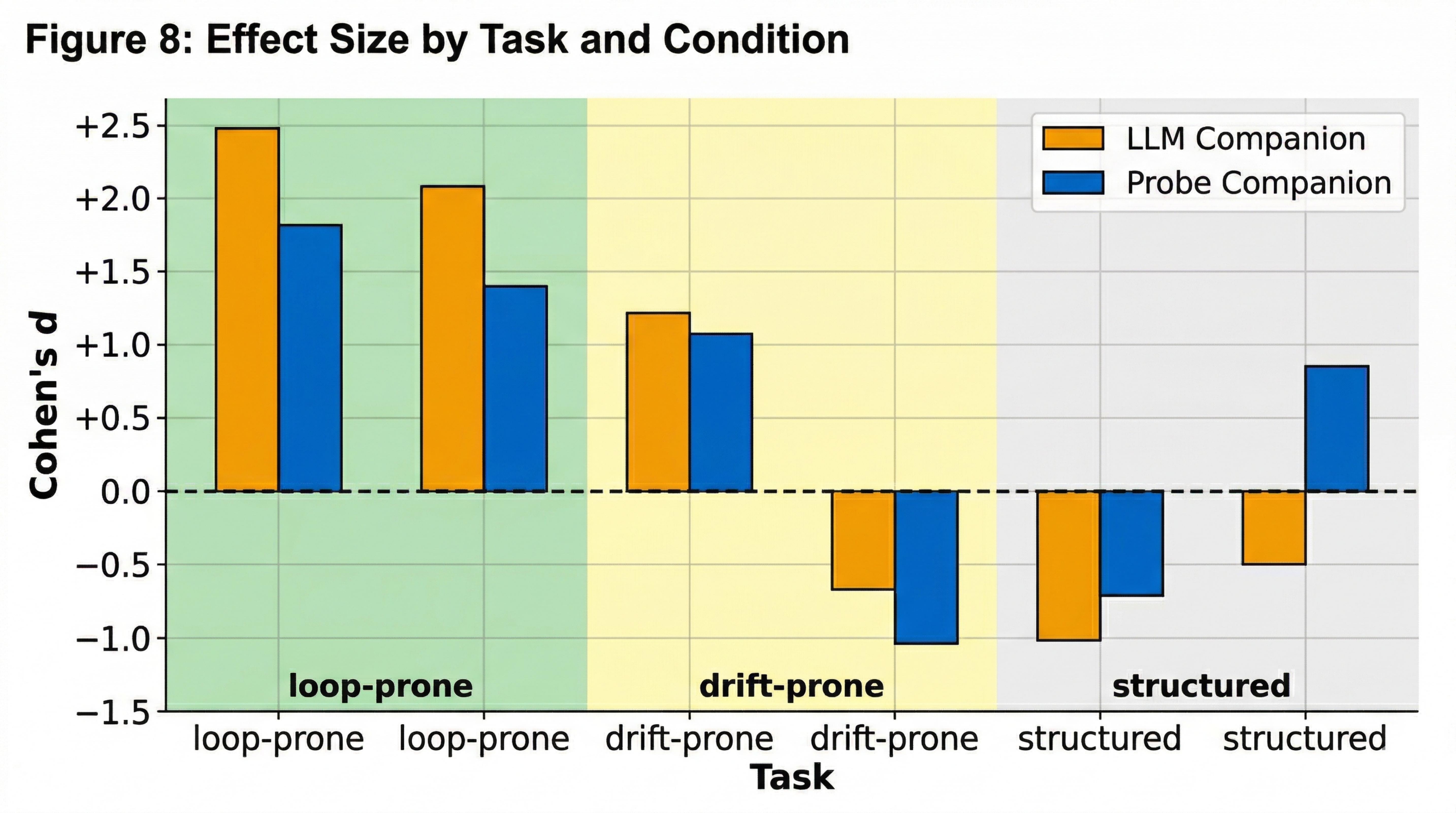}
    \caption{Effect sizes by task category revealing clear task-type dependency of companion effectiveness.}
    \label{fig:e2_by_task}
\end{figure}

\begin{figure}[htbp]
    \centering
    \includegraphics[width=0.85\linewidth]{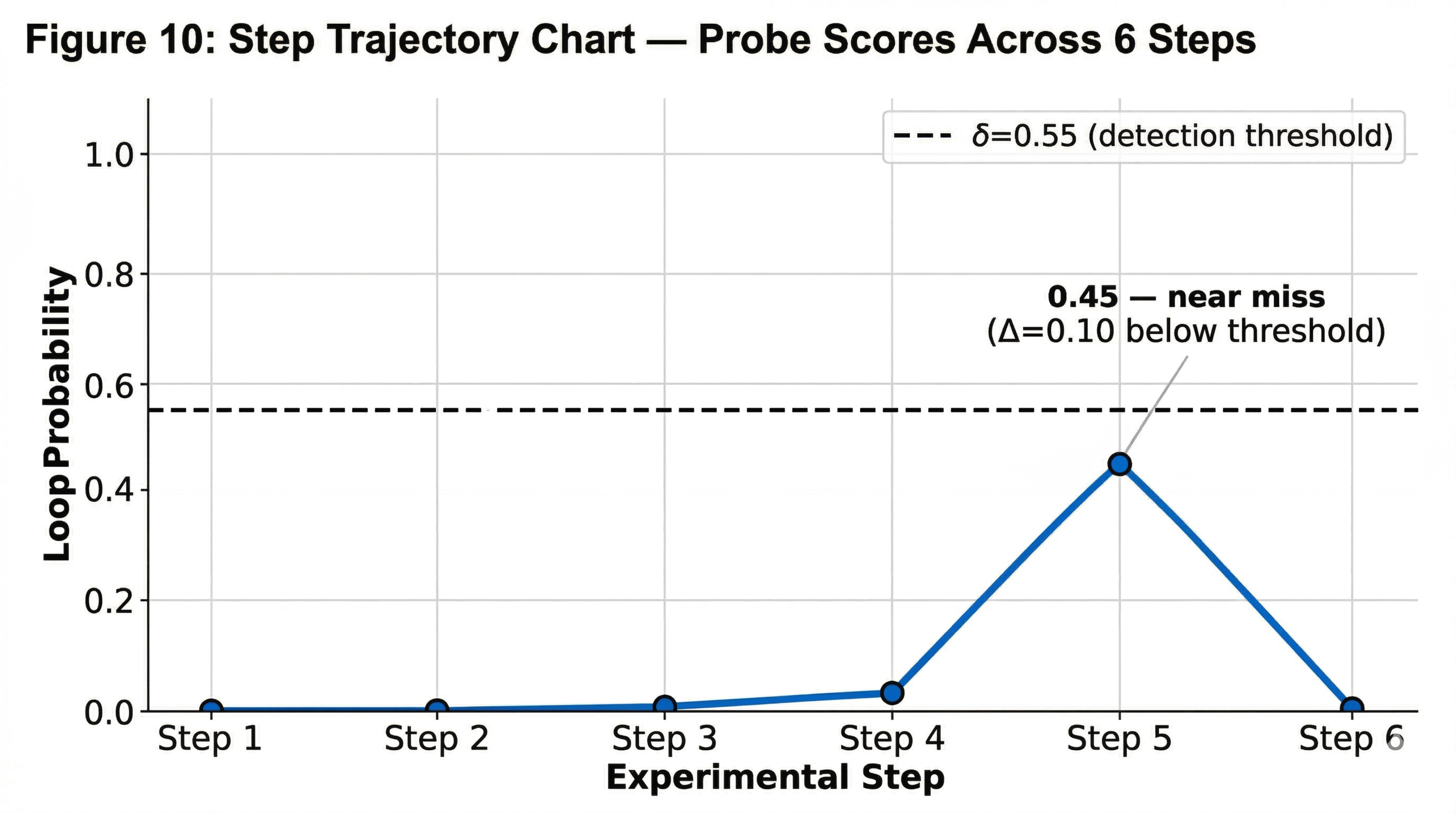}
    \caption{Probe degradation probability across reasoning steps showing signal detection and threshold sensitivity.}
    \label{fig:probe_trajectory}
\end{figure}

\begin{figure}[htbp]
    \centering
    \includegraphics[width=0.65\linewidth]{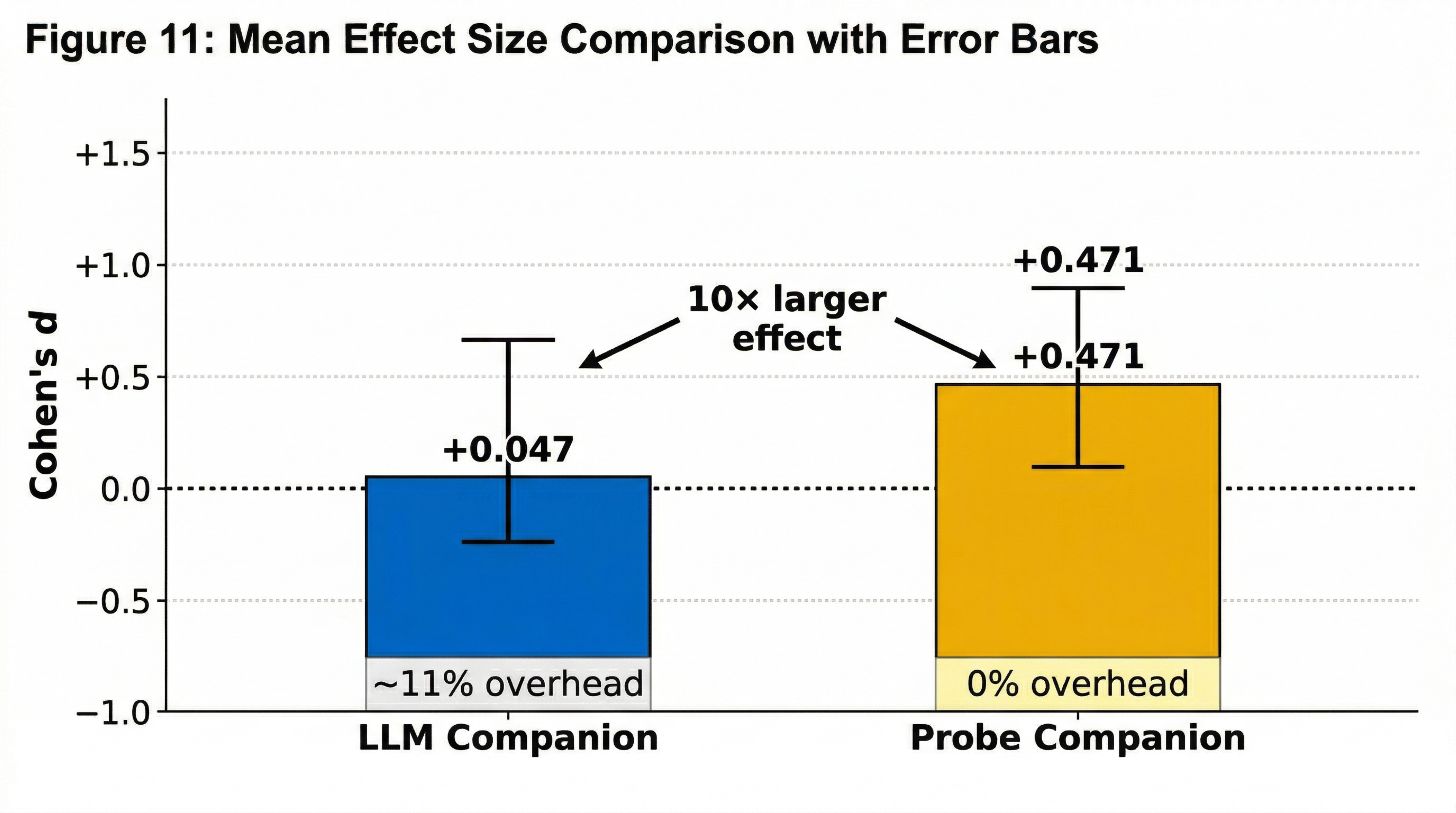}
    \caption{Mean effect size comparison demonstrating Probe Companion's superior performance at zero computational cost.}
    \label{fig:mean_effect_comparison}
\end{figure}

\begin{figure}[htbp]
    \centering
    \includegraphics[width=0.85\linewidth]{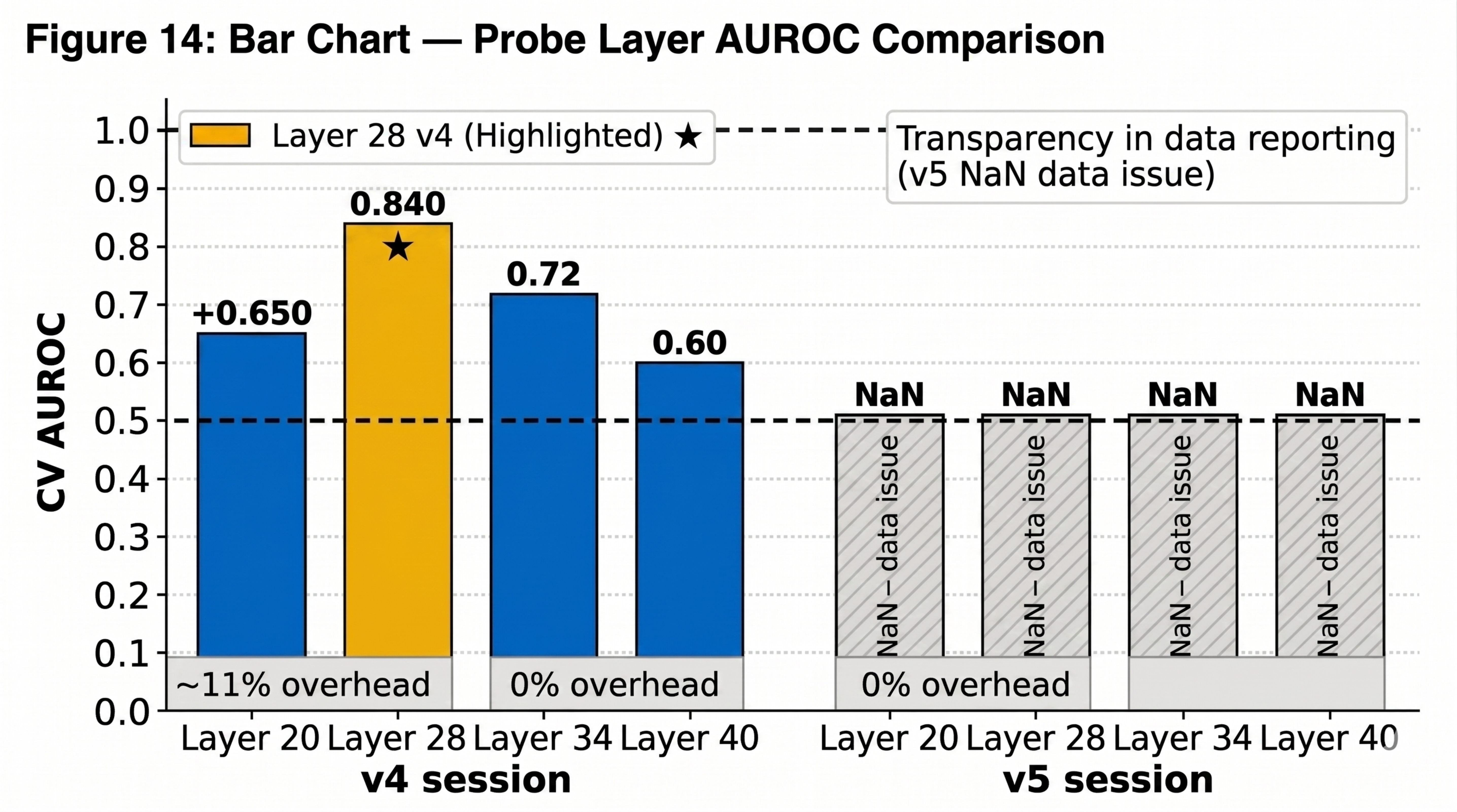}
    \caption{Probe layer selection with transparent disclosure of v5 data collection issues.}
    \label{fig:probe_layer_comparison}
\end{figure}

Figure~\ref{fig:e2_by_task} complements the per-task tables by showing how effect sizes cluster by task category. Figure~\ref{fig:probe_trajectory} provides an example probe-score trajectory relative to the intervention threshold, Figure~\ref{fig:mean_effect_comparison} summarizes the aggregate gap between the two companion types, and Figure~\ref{fig:probe_layer_comparison} visualizes the layer-selection results discussed alongside Table~\ref{tab:probe_layers}.

\begin{table}[htbp]
\centering
\begin{tabular}{|l|l|c|c|l|}
\hline
\textbf{Task} & \textbf{Category} & \textbf{LLM d} & \textbf{Probe d} & \textbf{Winner} \\
\hline
Liar Paradox & Loop-prone & +0.71 & \textbf{+2.04} & Probe \\
Ship of Theseus & Loop-prone & -0.82 & -0.82 & Baseline \\
Startup Design & Drift-prone & 0.00 & 0.00 & Baseline \\
Consciousness & Drift-prone & +0.43 & \textbf{+0.87} & Probe \\
DB Decision & Structured & -0.82 & 0.00 & Baseline \\
Algorithm Design & Structured & \textbf{+0.77} & +0.73 & LLM \\
\hline
\end{tabular}
\caption{E2: Per-task effect sizes (Cohen's d) showing task-type dependency of companion effectiveness.}
\label{tab:e2_per_task}
\end{table}

\begin{table}[htbp]
\centering
\begin{tabular}{|l|c|c|c|}
\hline
\textbf{Condition} & \textbf{Mean d (score)} & \textbf{Mean d (repetition)} & \textbf{Overhead} \\
\hline
LLM Companion & +0.047 ± 0.658 & +0.014 ± 0.723 & ~11\% \\
Probe Companion & \textbf{+0.471 ± 0.893} & \textbf{-0.646 ± 1.002} & \textbf{0\%} \\
Baseline wins & 3/6 tasks & — & 0\% \\
\hline
\end{tabular}
\caption{E2: Aggregate effect sizes with standard deviation showing Probe Companion's superior performance at zero overhead.}
\label{tab:e2_aggregate}
\end{table}

\begin{table}[htbp]
\centering
\begin{tabular}{|l|c|c|c|l|}
\hline
\textbf{Layer} & \textbf{Train AUROC} & \textbf{CV AUROC} & \textbf{Examples} & \textbf{Data Source} \\
\hline
Layer 20 & 1.000 & 0.780 & 35 & v4 session \\
\textbf{Layer 28} & \textbf{1.000} & \textbf{0.840 $\star$} & \textbf{35} & \textbf{v4 session} \\
Layer 34 & 1.000 & 0.720 & 35 & v4 session \\
Layer 40 & 1.000 & 0.720 & 35 & v4 session \\
Layer 20 & 1.000 & NaN* & 32 & v5 session \\
Layer 28 & 1.000 & NaN* & 32 & v5 session \\
\hline
\multicolumn{5}{l}{*NaN due to only 3 DEGRADED examples — class absent in some CV folds} \\
\end{tabular}
\caption{Probe layer selection results showing Layer 28 as optimal with transparent reporting of v5 data issues.}
\label{tab:probe_layers}
\end{table}

\textbf{Probe Training Data Disclosure}: v5 collected only 3 DEGRADED examples (9.4\%), causing CV AUC = NaN. Valid result is v4 (AUC 0.840, 10 DEGRADED / 25 ON\_TRACK). Both reported transparently.

\subsection{Experiment 3: Small Model Scale Boundary (E3)}

\begin{figure}[htbp]
    \centering
    \includegraphics[width=0.85\linewidth]{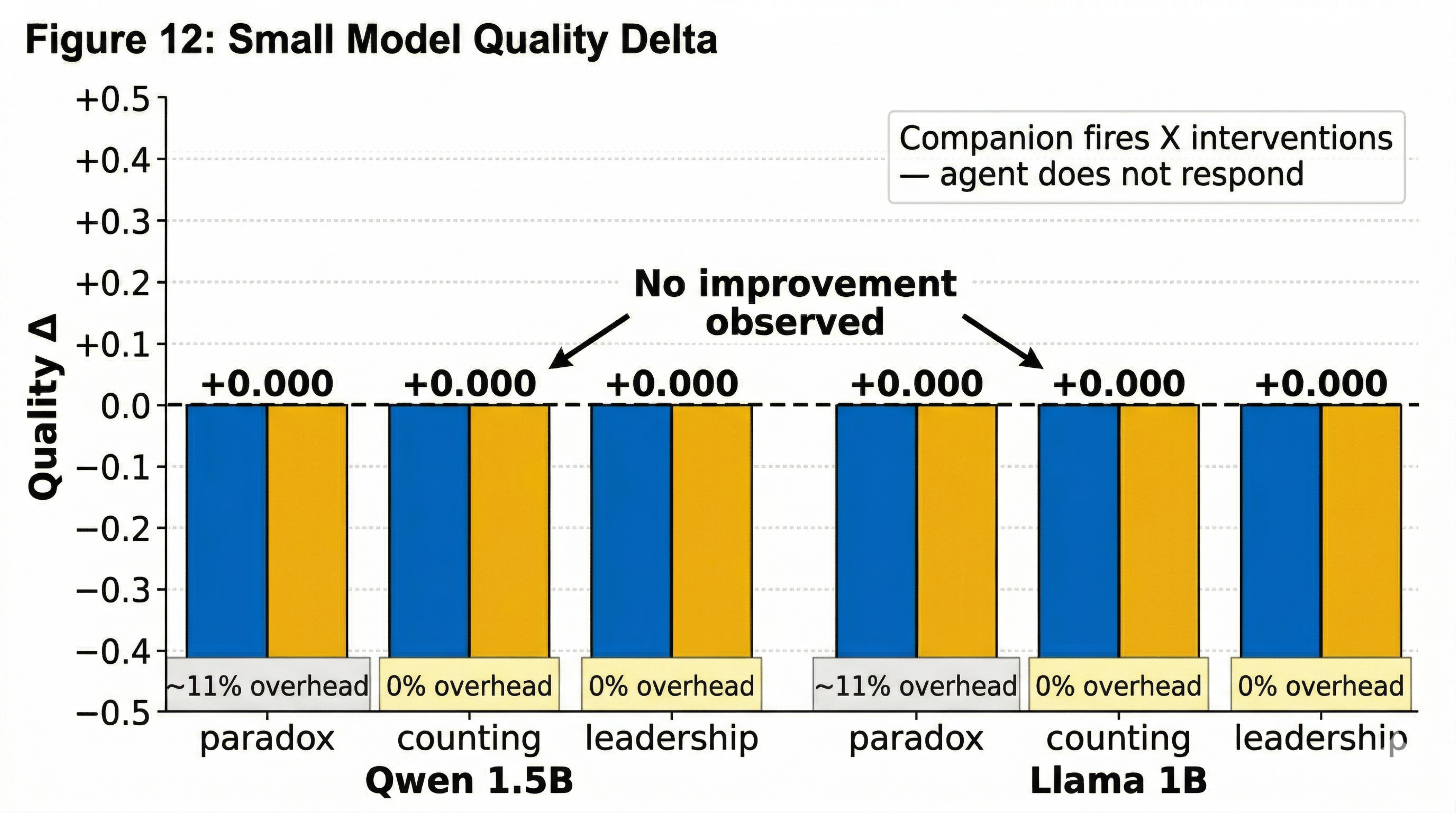}
\caption{Small-model results showing zero improvement in the study's quality proxy across all conditions, suggesting a scale boundary below Gemma 4 E4B.}
    \label{fig:small_model_results}
\end{figure}

Figure~\ref{fig:small_model_results} provides a compact view of the small-model quality results, making the absence of observed quality improvement across both model families visually explicit.

\begin{table}[htbp]
\centering
\begin{tabular}{|l|c|c|c|c|c|}
\hline
\textbf{Model} & \textbf{Params} & \textbf{Quality $\Delta$} & \textbf{Repetition $\Delta$} & \textbf{Avg Interventions} & \textbf{Tasks Improved} \\
\hline
Qwen 2.5 1.5B & 1.5B & \textbf{0.000} & +0.006 & 1.2 & 0/3 \\
Llama 3.2 1B & 1.0B & \textbf{0.000} & -0.121 & 0.8 & 0/3 \\
\textbf{Overall} & — & \textbf{0.000} & -0.057 & 1.0 & \textbf{0/6} \\
\hline
\end{tabular}
\caption{E3: Small-model results showing zero improvement in the study's quality proxy despite intervention firing, suggesting a possible scale boundary.}
\label{tab:e3_small_models}
\end{table}

\subsection{Task-Type Analysis}
\textit{Key emergent finding}

The most practically useful finding is that companion effectiveness depends strongly on task type in the current experiments. Table~\ref{tab:task_type_analysis} quantifies this relationship.

\begin{figure}[htbp]
    \centering
    \includegraphics[width=0.65\linewidth]{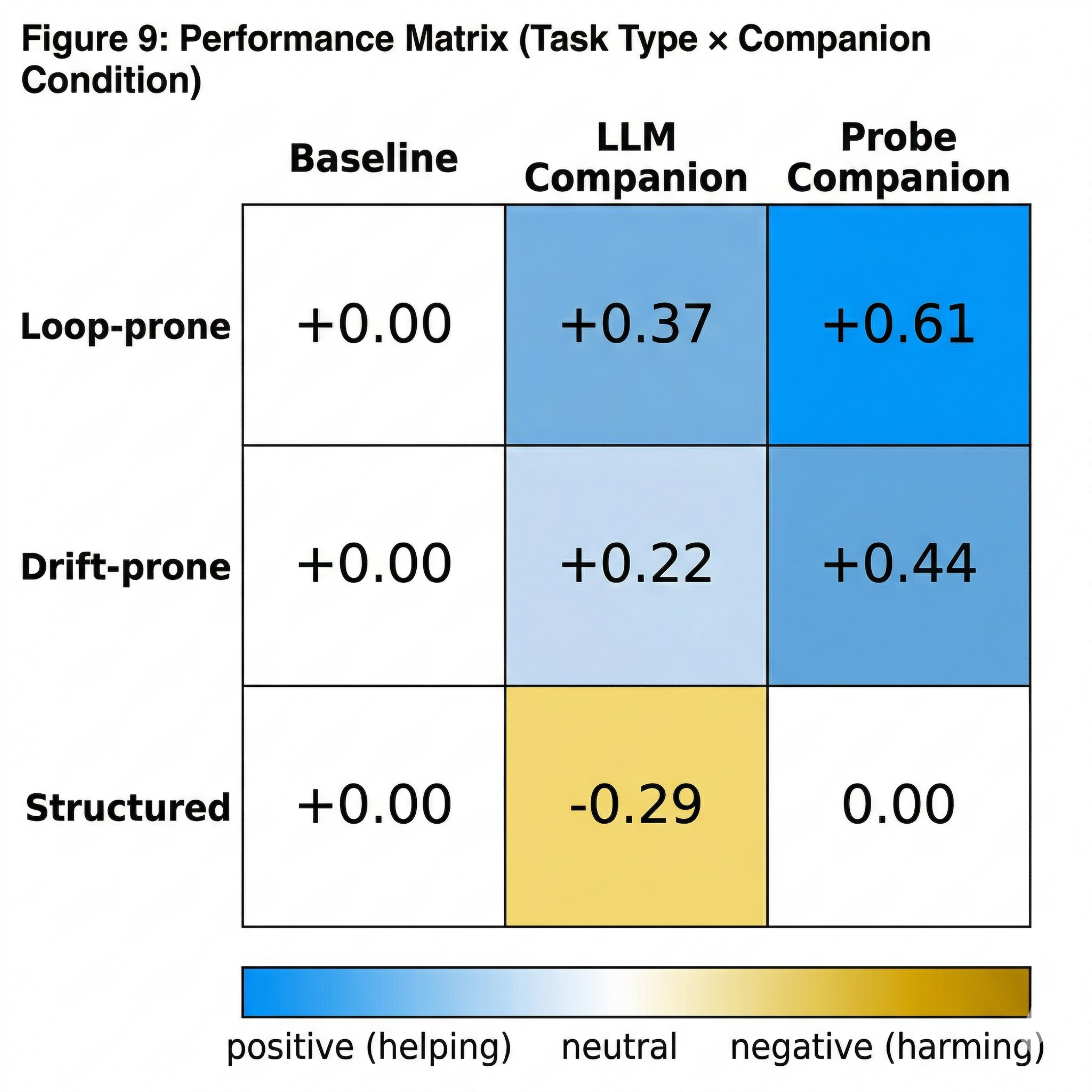}
    \caption{Performance heatmap revealing task-type specific deployment zones for companion activation.}
    \label{fig:performance_matrix}
\end{figure}

\begin{figure}[htbp]
    \centering
    \includegraphics[width=0.95\linewidth]{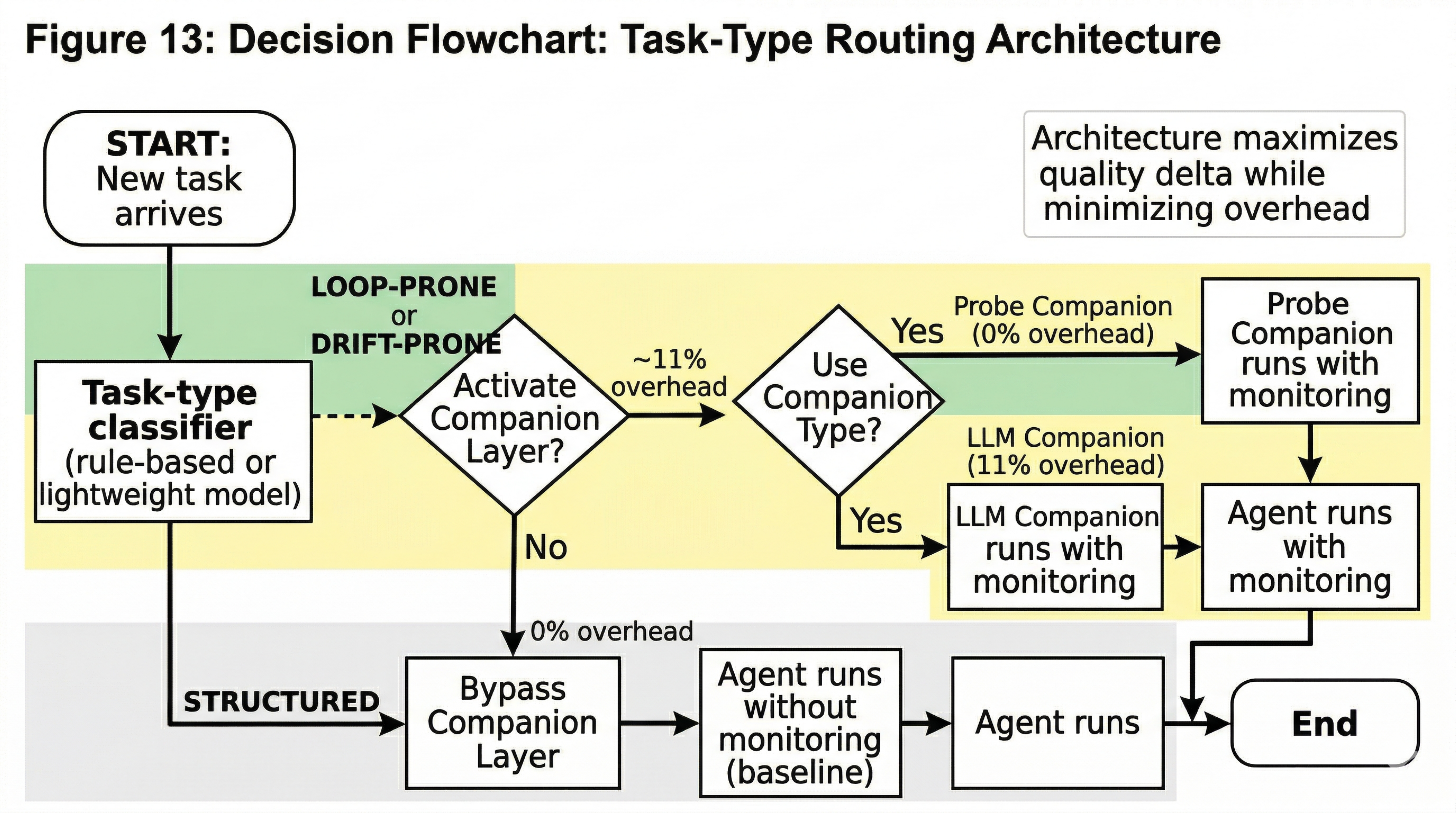}
    \caption{Proposed task-type routing architecture translating research findings into practical deployment strategy.}
    \label{fig:task_routing}
\end{figure}

Figure~\ref{fig:performance_matrix} summarizes this task-type pattern as a performance matrix, and Figure~\ref{fig:task_routing} translates the same empirical pattern into a tentative selective-routing architecture.

\begin{table}[htbp]
\centering
\begin{tabular}{|l|l|c|c|l|}
\hline
\textbf{Task Category} & \textbf{Tasks} & \textbf{LLM mean d} & \textbf{Probe mean d} & \textbf{Recommendation} \\
\hline
Loop-prone & Liar Paradox, Ship of Theseus & +0.37 & \textbf{+0.61} & Consider companion \\
Drift-prone & Consciousness, Startup & +0.22 & \textbf{+0.44} & Consider companion \\
Structured & DB Decision, Algorithm & -0.29 & 0.00 & Prefer baseline \\
\hline
\end{tabular}
\caption{Task-type analysis showing tentative deployment heuristics based on mean effect sizes by category.}
\label{tab:task_type_analysis}
\end{table}

\subsection{Presence Effect}
\textit{Observation}

In multiple experiments, companions fired zero interventions yet achieved quality similar to or better than baseline. We refer to this tentatively as a \textbf{presence effect} and outline three non-exclusive explanations: (1) observer bias in self-assessment, (2) stochastic variation in sampling, and (3) latent priming from companion initialization. At present this remains a hypothesis-generating observation rather than a validated mechanism.

\section{Discussion}

\subsection{Key Findings}

\begin{enumerate}
\item \textbf{Feasibility Supported}: Both companions can reduce repetition in some settings, and the probe achieves this at zero measured inference overhead within the present implementation.
\item \textbf{Task-type Matters}: Companion effectiveness varies more by task category than by a simple ``companion vs no companion'' framing.
\item \textbf{Small-model Limits are Plausible}: The small-model study suggests that 1B--1.5B models may not reliably act on guidance despite successful intervention firing.
\item \textbf{Probe Results are Promising but Preliminary}: The observed effect size advantage (d=+0.471 vs d=+0.047) is encouraging, but should be interpreted in light of proxy labels, small sample size, and in-domain evaluation.
\item \textbf{Selective Deployment is Better Motivated than Universal Deployment}: Negative effects on some structured tasks suggest companions should be activated conservatively and conditionally.
\end{enumerate}

\subsection{Limitations and Scope}

This feasibility study acknowledges significant methodological constraints (see Table~\ref{tab:limitations}) that limit generalizability. The v5 probe training yielded only 3 DEGRADED examples (9.4\%), causing NaN cross-validation AUROC. The strongest probe result therefore derives from a small earlier run with 35 proxy-labeled examples and AUROC 0.840. While this is enough to motivate the approach, it is not sufficient for robust deployment claims.

\subsection{Implications for Production Deployment}

The task-type dependency finding suggests a \textbf{task-type routing architecture} where companions activate only for loop-prone and drift-prone tasks while bypassing structured tasks. At present, this should be read as a deployment heuristic motivated by the observed pattern rather than a validated policy. If the pattern holds in larger studies, selective deployment could improve the accuracy-efficiency trade-off:

\textbf{Loop-prone/Drift-prone tasks} $\rightarrow$ Activate companion (Probe preferred for 0\% overhead)

\textbf{Structured tasks} $\rightarrow$ Baseline execution (companion causes harm)

This approach transforms the unexpected task-type finding into practical deployment guidance, potentially enabling widespread adoption in production agent systems.

\section{Future Work}

\begin{figure}[htbp]
    \centering
    \includegraphics[width=0.85\linewidth]{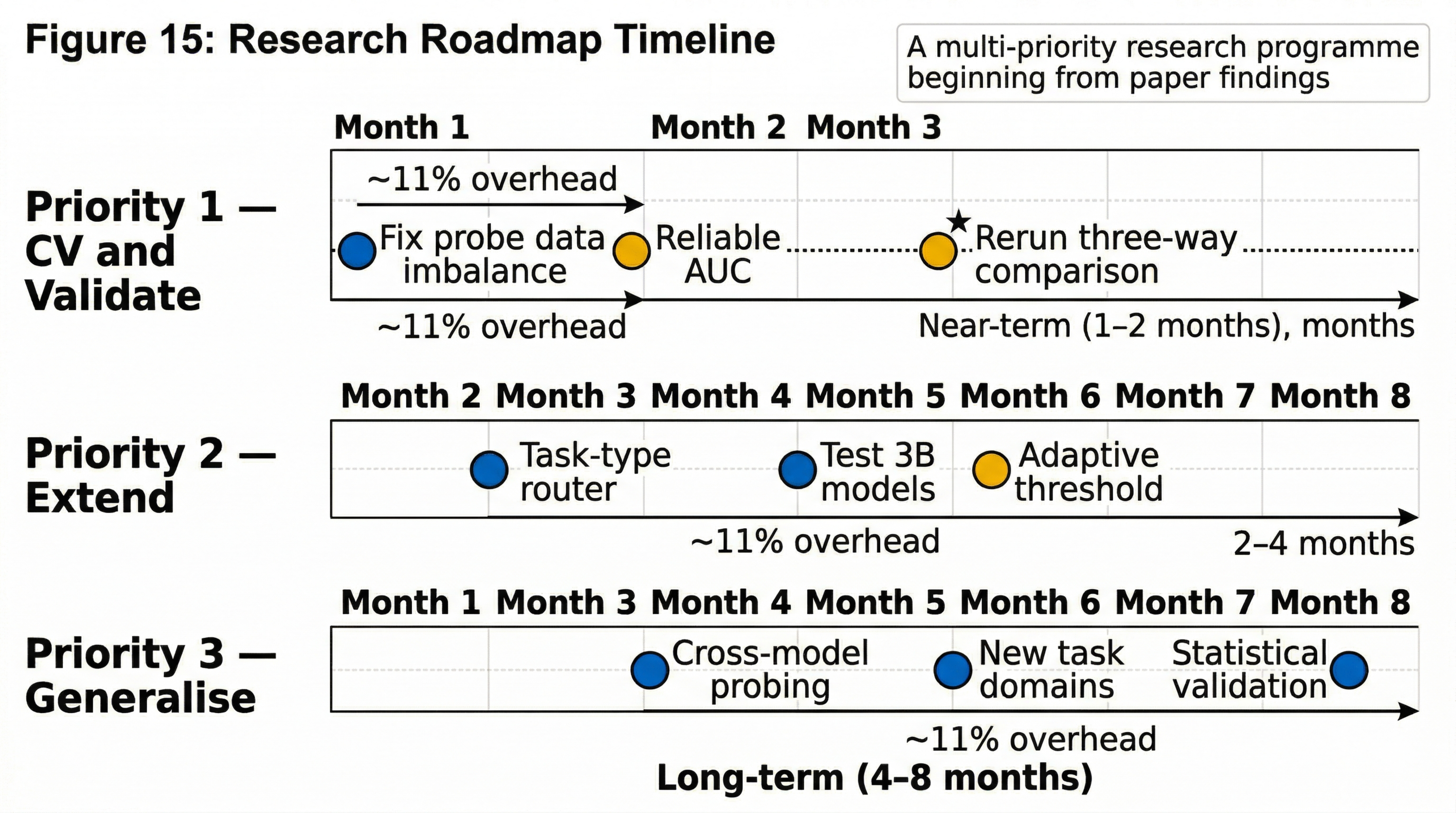}
    \caption{Three-priority research roadmap positioning this work as foundation for systematic companion development.}
    \label{fig:research_roadmap}
\end{figure}

This feasibility study establishes a foundation for rigorous investigation across multiple dimensions. Figure~\ref{fig:research_roadmap} outlines the comprehensive research program.

\subsection{Priority 1: Methodological Improvements (Blocking Current Limitations)}

\textbf{Fix Probe Data Imbalance}: Current 3 DEGRADED examples (9.4\%) in v5 are insufficient. Target collection of 200+ examples per class through harder tasks or longer reasoning sequences to ensure robust cross-validation.

\textbf{Fix Three-way Comparison}: Resolve KeyError in statistical comparison framework to enable clean baseline vs LLM vs Probe evaluation with proper effect size reporting.

\textbf{External Judge Model}: Replace self-referential quality assessment with external model or human annotation to eliminate circular bias in evaluation.

\subsection{Priority 2: Extending the Core Finding (Task-Type Specificity)}

\textbf{Task-Type Routing Implementation}: Develop and validate automated task classification to enable selective companion activation based on loop-prone/drift-prone/structured categorization.

\textbf{Test 3B Model Scale}: Validate Qwen 2.5 3B and Llama 3.2 3B as minimum viable scale for companion effectiveness, bridging the gap between 1.5B (ineffective) and 4.5B (effective).

\textbf{Adaptive Threshold Calibration}: Develop task-complexity and confidence-aware threshold adjustment to optimize intervention precision while reducing false positive rates.

\subsection{Priority 3: Generalization and Production Scale}

\textbf{Cross-Model Probe Transfer}: Train and evaluate probe classifiers across multiple architectures (Llama 3, Qwen 2.5, Claude) to establish generalizability of hidden state monitoring.

\textbf{New Task Domains}: Extend evaluation beyond philosophical reasoning to code generation, tool usage, and structured problem-solving to validate broader applicability.

\textbf{Statistical Significance Framework}: Implement multi-run experimental design with proper confidence intervals, effect sizes, and significance testing for rigorous evaluation.

\textbf{Production Integration}: Direct integration with agent frameworks (LangGraph, AutoGen, OpenHands) to enable real-world validation and deployment.

\section{Conclusion}

To our knowledge, this work is an early demonstration of zero-overhead cognitive state monitoring for LLM agents through hidden state analysis. The results support the feasibility of lightweight semantic monitoring, while also showing that task-type dependency is likely to be a critical constraint on deployment.

\textbf{What is Supported by the Current Evidence}: In the present experiments, Probe-based monitoring achieved mean effect size +0.471 with zero measured inference overhead, and the strongest small probe study achieved cross-validated AUROC 0.840 on a proxy-labeled dataset. Across tasks, companion benefits were concentrated in loop-prone and open-ended settings, while structured tasks often saw neutral or negative effects.

\textbf{What Appears Novel}: The probe approach leverages hidden states extracted during existing forward passes without additional model calls, making semantic monitoring nearly free at inference time. The observed task-type specificity also reframes companion systems as selective tools rather than universally helpful supervisors.

\textbf{What is Not Yet Proven}: Generalization across models, statistical significance with larger datasets, held-out-task probe performance, and production-scale validation remain open questions. The acknowledged methodological limitations — small proxy-labeled datasets, in-domain evaluation, single-model emphasis, and self-referential judging — require systematic treatment in future work.

\textbf{Foundation for Future Work}: This feasibility study establishes architectural feasibility and provides a concrete roadmap for stronger follow-up work. The modular design, transparent limitation reporting, and selective-routing hypothesis together create a useful base for larger, better-controlled experiments.

As LLM agents become increasingly central to automated reasoning workflows, the zero-overhead monitoring principles and task-type sensitivity findings demonstrated here provide a pathway toward more reliable, efficient, and contextually appropriate agent supervision.

\bibliographystyle{unsrtnat}

\textbf{Code and Data Availability}: Implementation code and experimental data will be made publicly available to enable reproducibility and community extension.

\end{document}